\documentclass[a4paper, 11pt, numbers=enddot, ngerman]{scrartcl}
\usepackage[T1]{fontenc}
\usepackage[utf8]{inputenc}
\usepackage{xkvltxp}
\usepackage[main=english,ngerman,english]{babel}
\usepackage[babel]{csquotes}
\usepackage[sort]{cite}

\usepackage{algorithm,algpseudocode}
\floatname{algorithm}{Algorithm}

\usepackage[acronym, nonumberlist, nopostdot]{glossaries}
\newacronym{ahc}{AHC}{Agglomerative Hierarchical Clustering}
\newacronym{lcl}{LCL}{Lane Change Left}
\newacronym{lk}{LK}{Lane Keeping}
\newacronym{lcr}{LCR}{Lane Change Right}
\newacronym{voi}{VOI}{Vehicle Of Interest}
\newacronym{svm}{SVM}{Support Vector Machine}
\newacronym{gda}{GDA}{Gaussian Discriminant Analysis}
\newacronym{gmm}{GMM}{Gaussian Mixture Model}
\newacronym{bdt}{BDT}{Boosted Decision Trees}

\usepackage{calrsfs}
\DeclareMathAlphabet{\pazocal}{OMS}{zplm}{m}{n}

\usepackage{bm}

\usepackage{import}

\usepackage{etoolbox}

\usepackage[uppercase]{titlesec}
\usepackage{wordlike}

\usepackage[parfill]{parskip}

\usepackage{setspace}
\titleformat{\section}{\normalfont\fontsize{12}{0}\bfseries}{\thesection.}{1em}{}
\titlespacing\section{0pt}{0pt minus 0cm}{0pt minus 1cm}
\titleformat{\subsection}{\normalfont\fontsize{12}{0}\bfseries}{\thesubsection.}{1em}{}
\titlespacing\section{0pt}{0pt minus 0cm}{0pt minus 0cm}
\titleformat{\subsubsection}{\normalfont\fontsize{12}{0}\bfseries}{\thesubsubsection.}{1em}{}
\titlespacing\section{0pt}{0pt minus 0cm}{0pt minus 0cm}

\usepackage[bottom]{footmisc}  

\usepackage{amsmath}
\usepackage{amssymb}
\usepackage{subcaption}

\usepackage{url}

\makeatletter
\g@addto@macro\@floatboxreset\centering
\makeatother

\usepackage{enumitem}
\setlist[itemize]{noitemsep, topsep=0pt}

\makeatletter
\renewcommand{\section}{\@startsection{section}{1}{\z@}%
	{-3.5ex \@plus -1ex \@minus -.2ex}%
	{2.3ex \@plus .2ex}%
	{\normalfont\Large\scshape\bfseries}}

\renewcommand{\subsection}{\@startsection{subsection}{2}{\z@}%
	{-3.25ex\@plus -1ex \@minus -.2ex}%
	{1.5ex \@plus .2ex}%
	{\normalfont\large\scshape\bfseries}}

\renewcommand{\subsubsection}{\@startsection{subsubsection}{2}{\z@}%
	{-3.25ex\@plus -1ex \@minus -.2ex}%
	{1.5ex \@plus .2ex}%
	{\normalfont\normalsize\scshape\bfseries}}
\makeatother
\usepackage{pgfplots} 
\pgfplotsset{compat=newest} 
\pgfplotsset{plot coordinates/math parser=false} 

\usepackage{tikz, tikzscale}
\usetikzlibrary{patterns}

\usepackage{multirow}
\usepackage{booktabs}
\usepackage{color}
\usepackage{xspace}
\usepackage{icomma}
\usepackage{expdlist}

\newcommand{\comment}[1]{}

\begin{document}

\textbf{\Large Prediction of Highway Lane Changes Based on Prototype Trajectories}  \bigskip


{\large \textbf{David Augustin} M.Sc., Marius Hofmann B.Sc., Opel Automobile GmbH; Prof. Dr.-Ing. Ulrich Konigorski, TU Darmstadt} \\ \bigskip

\vspace{-1mm}
\textbf{Abstract} \\
The vision of automated driving is to increase both road safety and efficiency, while offering passengers a convenient travel experience. This requires that autonomous systems correctly estimate the current traffic scene and its likely evolution. In highway scenarios early recognition of cut-in maneuvers is essential for risk-aware maneuver planning.
In this paper, a statistical approach is proposed, which advantageously utilizes a set of prototypical lane change trajectories to realize both early maneuver detection and uncertainty-aware trajectory prediction for traffic participants. Generation of prototype trajectories from real traffic data is accomplished by \glsdesc{ahc}. During clustering, the alignment of the cluster prototypes to each other is optimized and the cohesion of the resulting prototype is limited when two clusters merge. In the prediction stage, the similarity of observed vehicle motion and typical lane change patterns in the data base is evaluated to construct a set of significant features for maneuver classification via \glsdesc{bdt}. The future trajectory is predicted combining typical lane change realizations in a mixture model. B-splines based trajectory adaptations  guarantee continuity during transition from actually observed to predicted vehicle states. Quantitative evaluation results demonstrate the proposed concept's improved performance for both maneuver and trajectory prediction compared to a previously implemented reference approach.

\textbf{Keywords} maneuver detection, trajectory prediction, motion pattern recognition, highly automated driving, boosted decision trees


\section{Introduction}

The development of automated driving functions is a central activity of industry and science. Currently, first assistance systems combining both longitudinal and lateral guidance are introduced on the market.
In order to autonomously drive in public traffic these systems need to correctly evaluate the current traffic scene and its likely evolution. This includes estimating the driving intentions of traffic participants, detecting maneuvers being performed, and predicting future motion of surrounding traffic. All these tasks are non-deterministic and uncertainties in maneuver and motion estimation need to be considered.
In this contribution, highly automated driving (HAD) on highways is addressed with the focus on early lane change maneuver detection and uncertainty-aware trajectory prediction.
Before outlining the basic concepts of the proposed approach in Subsection \ref{ProposedApproach} an overview of related work in the field of motion prediction is provided.

\subsection{Related Work} \label{RelatedWork}

Motion prediction is a research area since decades. The survey articles \cite{lefevre2014survey} and \cite{Liebner.2015} provide overviews about a variety of published approaches and fields of application related to road transport. Additionally, both provide suggestions on how motion prediction approaches can be categorized.\\
Regarding driver intention recognition, a large number of approaches do assess the traffic environment on an abstract level while neglecting the prediction of physical motion which is crucial for criticality assessment and trajectory planning. The opposite concept relying purely on physically-based motion prediction is only valid for a limited time horizon of less than a second \cite{lefevre2014survey}. Approaches that realize both motion detection and trajectory prediction can be classified as methods of pattern recognition and methods fusing dynamic motion models with behavior descriptions \cite{schreier2016integrated}. Methods of pattern recognition do not only consider the current state of an object but also its history. These states are compared to a database of previously generated motion patterns, each linked to a specific maneuver type. In scenarios like intersections, in which the road topology predefines typical trajectories, the patterns can be directly created from map data \cite{wyder2015bayesian}. In \cite{houenou2013vehicle} trajectory prediction for lane changes is realized combining a physical motion model with artificially created motion patterns. At each time step a new set of motion patterns is generated based on the current vehicle state, the road parameters, and the detected maneuver. Another possibility for pattern creation is to analyze real traffic data. 
The approach \cite{bahram2016combined} utilizes a set of recorded highway lane change trajectories as prototypes in a Bayesian network classifier for maneuver detection. Clustering methods are not applied \cite{Bahram.2017}. Motion prediction for traffic participants is realized using multi-agent simulation.
In \cite{vasquez2004motion} a cluster-based technique that learns the typical motion patterns using pairwise clustering is designed for a visual pedestrian tracking system placed in an entry hall. For static environments like interior rooms of buildings or intersections it is fairly easy to define regions of start and end points of motion patterns compared to environments with varying topology like highways. 

\subsection{Proposed Approach} \label{ProposedApproach}

\begin{figure}[thpb]
	\centering
	\includegraphics[width=0.7\linewidth]{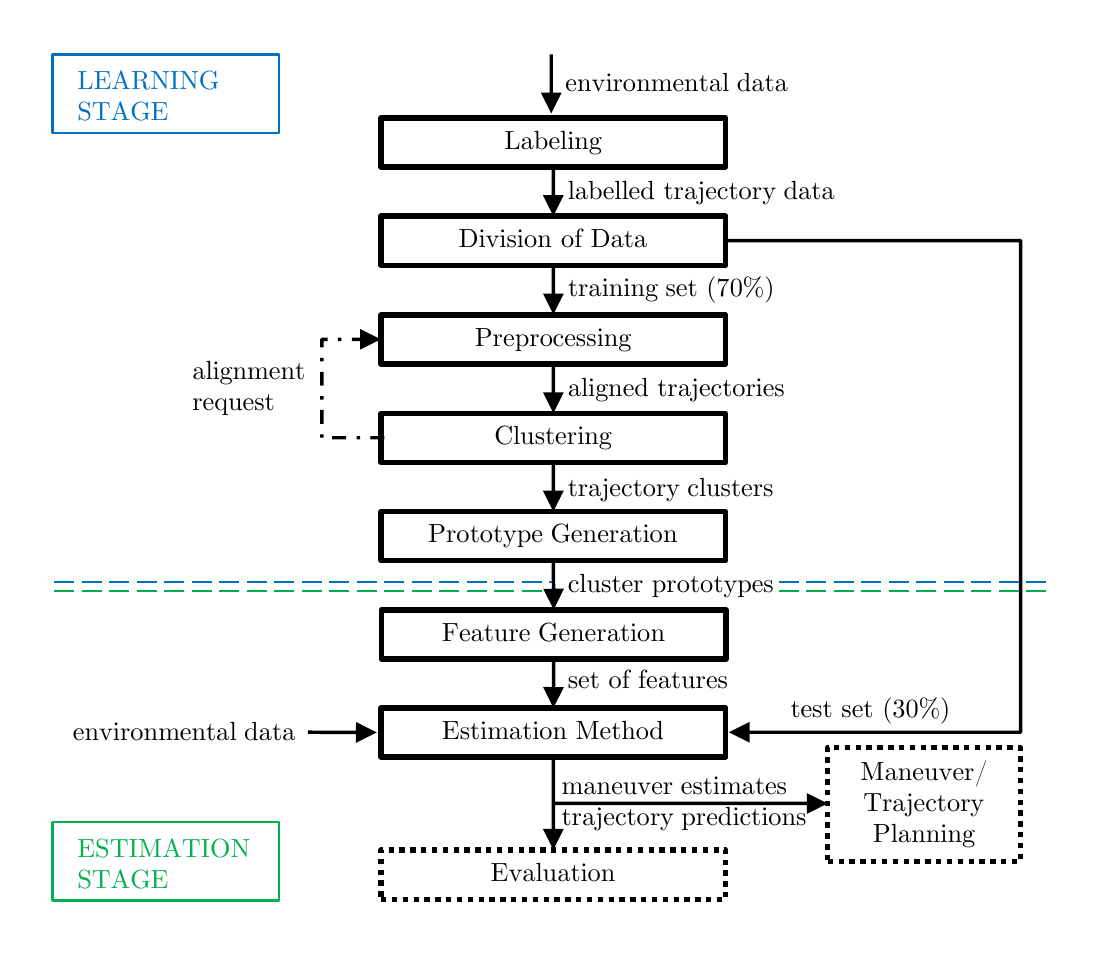} 
	\caption{Process diagram of the proposed approach. Maneuver and trajectory estimates can be used as inputs for HAD maneuver and trajectory planning.}
	\label{process_diagram}
\end{figure}

In this contribution we propose an extension to a previously published approach for lane change maneuver detection and uncertainty-aware trajectory prediction in highway scenarios \cite{Augustin.2018}. The approach is centered around the idea that the variety of lane change realizations can be clustered into typical motion patterns which support early maneuver recognition and trajectory prediction. Similar lane change courses are aggregated into groups, each represented by a prototype trajectory. Figure \ref{process_diagram} shows a schematic diagram of the proposed approach.\\
The objectives of learning stage are to produce a set of representative lane change trajectories, which can be utilized to formulate descriptive features for maneuver recognition, and to learn a convenient classification algorithm. Prototype trajectories are generated from recorded highway footage of traffic participants applying \gls{ahc} in combination with a preprocessing step optimizing the trajectory alignment. Additionally, constraints are introduced to improve cluster cohesion and define  the end condition for pairwise clustering. To reduce the diversity of lane change embodiments and thus limit the number of required representatives we categorize lane change trajectories solely by their lateral course. This enables maneuvers of different absolute velocity but similar lateral movement to be cumulated in the same cluster. Each lane change cluster is represented by lateral position and velocity prototypes. Details of typical motion pattern extraction from real traffic data are provided in Chapter \ref{motion_pattern_recognition}.\\
In the estimation stage, which is covered in Section \ref{maneuver_classification}, we buffer the latest states for each \gls{voi}. The vehicle state histories are referred to as partially observed trajectories. A distance value is computed for each pair of partial and prototype trajectory under optimal alignment for lateral position and velocity independently. For each lane change type (\gls{lcl} or \gls{lcr}) we select the pairs with minimal distances as best matches. The distance values of the best matches for \gls{lcl} and \gls{lcr} are utilized to construct features for maneuver detection. Additionally, the lateral offset of a vehicle within its lane and its lateral velocity is applied for maneuver inference. All these features are derived from motion states of each \gls{voi}, individually. Vehicle interdependencies or infrastructure information are not taken into account. This limits the possible prediction horizon because potential causes for lane change maneuvers are not considered. Maneuvers can only be detected during execution. On the other hand, approaches with lower abstraction level limit the requirements on environment sensing system (Section \ref{environment}), i.e. there is no need to detect the vehicle driving in front of a \gls{voi} in order to reason about its influence on driving intentions. Due to the choice of features the proposed approach can be categorized as maneuver-based driver intention estimation.\\
The extensions compared to our previous publication \cite{Augustin.2018} mainly lie in advances of prototype usage for both maneuver detection and maneuver-based trajectory prediction.
We train an ensemble of decision trees by adaptive boosting for maneuver classification and discuss its performance in comparison to the generative classifier \gls{gda}. The trajectories of the \glspl{voi} are predicted according to the estimated maneuvers.
In the case of lane changes, the future lateral motion is expressed as a weighted sum of all maneuver-related prototypes forming a \gls{gmm}. The weight of each prototype correlates with its proximity to the partially observed trajectory.
Initial offsets between the actual lateral position of a \gls{voi} and the trajectory prediction are compensated by applying a B-spline approximation. Longitudinal motion is covered independently by a (nearly) constant acceleration model.
Both classification and motion prediction are evaluated with a test set of recorded highway data. The results are discussed in Section \ref{results}.


\section{Environment Sensing} \label{environment}

The test vehicle is equipped with various radars and cameras for all-round object recognition. As we are interested in cut-in maneuver detection, we focus on vehicles located in the front of the test car.\\
\noindent A tracking system produces, for each sensed object, a local estimate of its state vector $\bm{X}^{\mathrm{(target)}}$:
\begin{align}
\bm{X}^{\mathrm{(target)}} = [x, y, \theta, v]^{T}
\end{align}
\noindent where $x$ and $y$ are local Cartesian coordinates, $\theta$ the relative heading angle, and $v$ the absolute velocity.
The camera system detects lane markings and approximates their course as cubic functions in the local coordinate system:
\begin{align}
y(x) = c_3 x^3+c_2 x^2 + c_1 x + c_0
\end{align}
\noindent with polynomial coefficients $c_3$, $c_2$, $c_1$, and $c_0$.\\
To be independent of the ego motion of the test vehicle, the states of all tracked objects are transformed into the Frenet frame. Centerlines of corresponding driving lanes of each traffic object are utilized as reference curves.	
Vehicle trajectories $\bm{\gamma}_{i}(t):[0, T_i] \rightarrow \mathbb{C}$ are expressed in the lane-relative configuration space $\mathbb{C}$ which is $[s, \dot{s}, d, \dot{d}]$ with $s$ being the traveled distance from the start of the maneuver, $\dot{s}$ the velocity in longitudinal direction, while $d$ and $\dot{d}$ being the lateral offset to the lane's centerline and the lateral velocity, respectively.
To cut down diversity of maneuver executions, the configuration space $\mathbb{C}$ is reduced to $[d,  \dot{d}]$ for cluster learning and representation. The lateral offset, $d$, can be derived using the minimal distance of the vehicle to its left, $d_\mathrm{l}$, and right lane markings, $d_\mathrm{r}$:
\begin{align}
d = \frac{1}{2} (d_\mathrm{l} + d_\mathrm{r})
\end{align} 
\noindent Thereby, the lane width is taken into account inherently. To reduce computational effort for the calculation of $d_\mathrm{l}$ and $d_\mathrm{r}$, lane marker polynomials are locally linearized. The maximal error of this approximation is below our measurement accuracy. The lateral velocity, $\dot{d}$, is derived by applying a low-pass filtered difference quotient. The longitudinal velocity, $\dot{s}$, is assumed to be equal to the vehicle's absolute velocity, $v$. The traveled distance, $s$, between two consecutive measurements is computed applying a constant velocity model \cite{Augustin.2018}.


\section{Motion Pattern Recognition} \label{motion_pattern_recognition}

Due to diversity of traffic situations and individual driving styles the courses of lane changes highly vary. Nevertheless, we assume that the variety of lane change realizations can be mapped to a finite number of typical motion patterns. Such a pattern represents a group of similar maneuver executions. As the number of clusters and the course of their representing trajectories are unknown, real highway traffic data is analyzed by a prototype-based clustering technique called \glsdesc{ahc}. Each maneuver type is analyzed separately. The structured highway environment limits vehicle maneuvers with respect to their lateral motion to \glsdesc{lcl}, \glsdesc{lk}, and \glsdesc{lcr}. Each data point in the data set is assigned to one maneuver type via automatic labeling.

\subsection{Labeling} \label{labeling}

Lane changes can easily be detected in a data set by searching for lane marker crossings. A more challenging aspect for reliable automatic labeling is to define start and end time of a maneuver. Before a vehicle passes the centerline of its starting lane we regard the motion as lane centering and assign it to the lane keeping maneuver. Also a possible overshoot on the target lane is treated as lane keeping. Additionally, it was determined that a maneuver bound is reached when the vehicle's absolute lateral velocity exceeds or undercuts a velocity threshold of $0.2 \, \mathrm{m/s}$. To improve the lateral velocity estimate, we approximate the vehicle's position course via B-spline curve and calculate lateral velocity, $\dot{d}$, via derivation.

\subsection{Preprocessing and Clustering} \label{clustering}

\begin{figure*}[t]
	\begin{subfigure}{0.45\linewidth}
		\centering
		\includegraphics[width=0.9\linewidth]{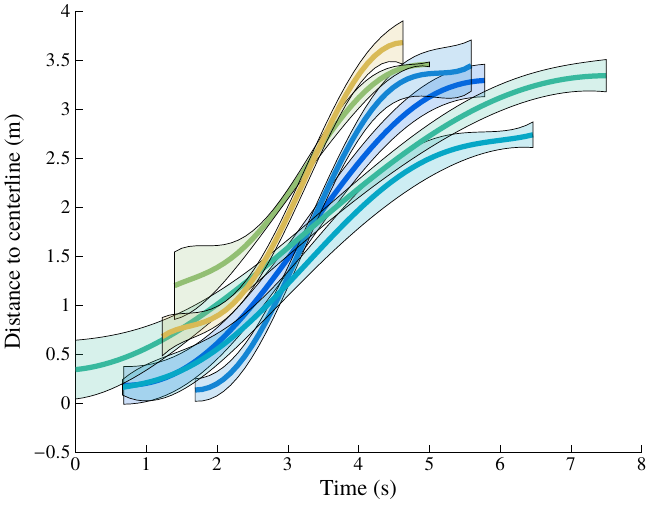} 
		\caption{Position prototypes for changes to the left lane.}
		\label{fig_clustering_maxSD_p_lcl}
	\end{subfigure}\hspace{0.2cm}%
	\begin{subfigure}{0.45\linewidth}
		\centering
		\includegraphics[width=0.9\linewidth]{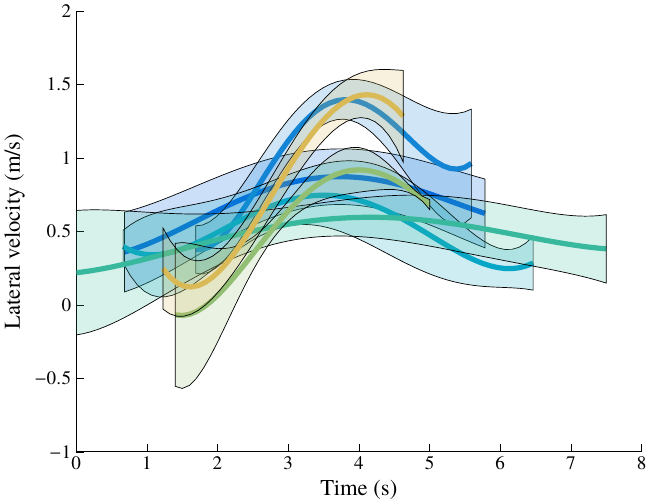}
		\caption{Velocity prototypes for changes to the left lane.}
		\label{fig_clustering_maxSD_v_lcl}
	\end{subfigure}%
	%
	\caption{Prototypical lane change trajectories created by \glsdesc{ahc}. For each lane change type six clusters are created from real traffic data.}
	\label{fig_clustering}
\end{figure*}

Objectives of traffic-related motion pattern recognition are to automatically assign similar vehicle trajectories to the same cluster and to compute a common representation for each group, which we denote as prototype trajectory. Due to formulation of goals prototype-based clustering methods, like \gls{ahc} or k-Means, are suitable solution approaches. Both methods were investigated and compared in \cite{Augustin.2018} with \gls{ahc} performing superior for highway scenarios.\\
\gls{ahc} is an algorithm iteratively merging the two closest clusters until only one single cluster exists or a termination condition is fulfilled \cite{tan2005introduction}. Initially, each cluster is formed by a unique trajectory of the training data set. When two clusters are merged, they are replaced by a common prototype and their similarity to all other cluster representatives must be reevaluated. Cluster representations consist of time-dependent mean, $\mu(t)$, and variance, $\sigma(t)$, values. The mean of one cluster, $C_m$, is calculated from its comprised trajectories by:
\begin{align}
\bm{\mu}_m(t) = \frac{1}{N_m} \sum \limits_{i=1}^{N_m} \bm{d}_i(t)
\end{align}
\noindent with $N_m$ denoting the total number of trajectories in the cluster. For prototype generation, \gls{ahc} is only applied to position trajectories, $\bm{d} \in \mathbb{D}$. $\mathbb{D}$ denotes the set of position trajectories relative to the centerline of the driving lane. For maneuver classification, it turned out to be beneficial to utilize both position and velocity prototypes, so in this case $\bm{d} \in \mathbb{D} \times \mathbb{V}$ with $\mathbb{V}$ denoting the set of velocity trajectories \cite{Augustin.2018}.\\
The scatter of cluster trajectories around the prototype mean is expressed via the variance
\begin{align}
\bm{\sigma}_m^{2}(t) = \frac{1}{N_m} \sum \limits_{i=1}^{N_m} (\bm{d}_i(t) - \bm{\mu}_m(t))^2
\end{align}
\noindent In order to express the relation between two trajectories a measure of proximity is required. We chose the average Euclidean distance between the two trajectories $\bm{d}_i, \bm{d}_j \in \mathbb{D}$  as dissimilarity measure:
\begin{align}
\delta(\bm{d}_i, \bm{d}_j) &= \left( \frac{1}{T} \int \limits_{t=t_\mathrm{min}}^{t_\mathrm{max}} (\bm{d}_i(t) - \bm{d}_j(t))^2dt \right)^{1/2} \\
t_\mathrm{min} &= \min(t_{0,i}, t_{0,j}) \notag \\
t_\mathrm{max} &= \max(t_{0,i} + T_i , t_{0,j} + T_j) \notag \\
T &= t_\mathrm{max} - t_\mathrm{min} \notag
\end{align}
\noindent where $t_{0,i}$, $t_{0,j}$, $T_i$, and $T_j$ are the starting times and the total maneuver durations of trajectories $\bm{d}_i$ and $\bm{d}_j$, respectively.	
For comparison of trajectories of unequal length, the shorter ones are extended by keeping their initial and/or final value. The factor $\frac{1}{T}$ averages the Euclidean distance to negate the influence of duration of the two trajectories being compared on their dissimilarity value.\newline
A major influence factor on the dissimilarity between two lane change trajectories is their mutual alignment \cite{Augustin.2018}, which in the case of highway trajectories is mainly determined by the applied labeling approach. In order to handle possible labeling inconsistencies we studied different approaches for active alignment of trajectories.	
Best results were achieved by an adjustment strategy minimizing the dissimilarity of each pair of trajectories by computing an individual optimal configuration \cite{Augustin.2018}. In \gls{ahc} the alignment of cluster representatives needs to be updated after each clustering iteration.\\
Figure \ref{fig_clustering} shows prototypical lane change motion patterns created using the presented \gls{ahc} approach.


\section{Maneuver Classification} \label{maneuver_classification}

Motion prediction for traffic participants can be split into two consecutive tasks. In the first step, the performed maneuvers need to be detected. Subsequently, the most probable future trajectories according to the maneuver estimate can be calculated. Maneuver detection is a typical classification problem. The highway topology limits the set of possible lateral maneuvers to $C = \mathrm{\{LCL, LK, LCR\}}$. We implemented maneuver inference based on a set of features which can be derived from the observed motion of individual vehicles. Vehicle interaction is not taken into account. This approach reduces computational and sensory  requirements. Additionally, not relying on strong assumptions about vehicle interdependencies increases robustness. On the other hand, maneuvers can only be recognized during their execution, thus limiting the possible prediction horizon. The applied features for maneuver detection of a \gls{voi} are lateral offset within its lane, lateral velocity and proximity of its motion path to prototype lane change trajectories in the database.
The proximity of a partially observed trajectory, $\bm{d}_p$, and a cluster prototype, $\bm{c}_m$, can be calculated via average Mahalanobis $L_1$ distance:
\begin{align}
\bm{\Delta}(\bm{d}_p, \bm{c}_m, \tau_m) &= \frac{1}{T_m} \int \limits_{-T_m}^{0} \left( \frac{( \bm{d}_p(t) - \mathbb{\mu}_m(t+\tau_m) )^2}{\bm{\sigma}_m^2(t+\tau_m)} \right)^{1/2} dt \\
T_m &= \min(T_\mathrm{buffer}, \tau_m) \nonumber
\end{align}
\noindent The evaluation period, $T_m$, is the minimum of buffered partial trajectory and prototype trajectory length. The cluster prototype is shifted by $\tau_m$ optimizing the trajectories' adjustment by minimizing their dissimilarity within the recent $0.5 \, \mathrm{s}$ \cite{Augustin.2018}. Compared to the Euclidean distance, utilized during clustering, the Mahalanobis distance measure takes the non-uniform uncertainty of the prototype course into account. Dissimilarities in regions of low variance have increased weight due to the term $\frac{1}{\bm{\sigma}_m^2}$.
Mahalanobis distance of position trajectories is denoted as $\bm{\Delta}_p$. The notation in case of velocity trajectories is $\bm{\Delta}_v$.\\
During feature generation a vehicle's partial trajectory is matched individually to all prototype trajectories in the database. For each pair $\{\bm{d}_p,\bm{c}_m\}$ the optimal shifting coefficient, $\tau_m$, is calculated minimizing $\bm{\Delta}_{p,[-0.5:0]}(\bm{d}_p,\bm{c}_m)$. The velocity-dependent distance $\bm{\Delta}_v(\bm{d}_p,\bm{c}_m)$ related to the same cluster is calculated for an equal value of $\tau_m$. As the best matching prototypes for \gls{lcl} and \gls{lcr} we select those minimizing the dissimilarity $\bm{\Delta}_p$ for the whole evaluation period. Their dissimilarity values are denoted $\bm{\Delta}_{p,\mathrm{LCL}}$, $\bm{\Delta}_{v,\mathrm{LCL}}$, $\bm{\Delta}_{p,\mathrm{LCR}}$ and $\bm{\Delta}_{v,\mathrm{LCR}}$.\\
In \cite{Augustin.2018} a Quadratic Gaussian Discriminant Analysis (\gls{gda}) classifier is applied for maneuver inference. In this contribution its performance is compared to \gls{bdt}, an ensemble of shallow decision trees sequentially trained by a boosting method, described in Section \ref{AdaBoost}.

\subsection{Boosting} \label{AdaBoost}

Boosting is a sequential method for consecutively fitting a set of basic learners to variations of the training data. After each iteration, the whole training set is evaluated \cite{Bishop.2006}. Misclassified data points are given greater weights when used to train the next classifier. Basic learners, also referred to as "weak learners" \cite{Murphy.2012}, typically have low classification performance when applied individually. In an ensemble of $N$ basic learners, each unit provides a class vote dependent on observed features. Those votes are weighted by $\alpha_n$ and combined like:
\begin{eqnarray}
y(\bm{f}) & = & \sum_{n=1}^N \alpha_n h_n(\bm{f})
\end{eqnarray}
\noindent with $h_n$ denoting the hypothesis of the $n$-th basic learner. We apply the Adaboost.M2 algorithm which is applicable in classification problems with multiple class instances \cite{freund1997decision}. In AdaBoost.M2 weighting coefficients $\alpha_n$ are derived from weighted pseudo-loss functions, providing more accurate basic classifiers with a greater weight.\\
As basic classifiers we use shallow decision trees limiting the maximum number of branch nodes to $15$. The ensemble classifier comprises $90$ basic decision trees. Treated individually, decisions are easily interpretable. They are not limited to a specific type of data. The flexibility of decision trees is extended by enlargement to ensemble methods at the expense of easy interpretability. AdaBoost application does not require hyperparameter training and showed best performance in an extensive empirical comparison \cite{caruana2006empirical}.


\section{Maneuver-based Trajectory Prediction} \label{motion_prediction}

\begin{figure*}[t]
	\centering
	\begin{subfigure}{0.45\linewidth}
		\centering
		\includegraphics[width=\linewidth]{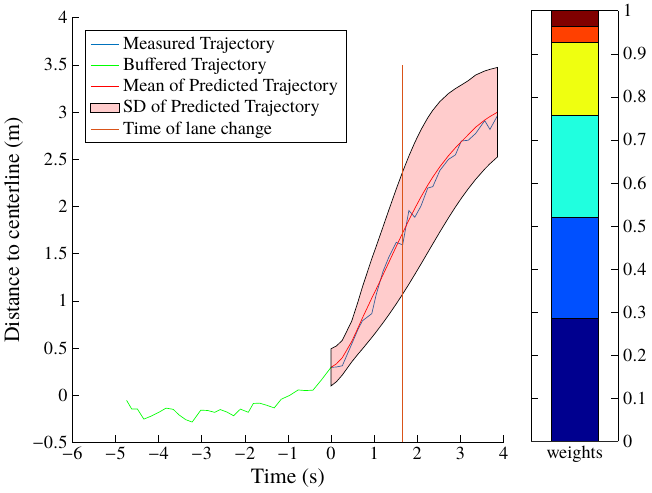} 
		\caption{Predicted trajectory 1.66s before the lane change.}
		\label{fig_trajectory_prediction_1_66}
	\end{subfigure}\hspace{5mm}
	\begin{subfigure}{0.45\linewidth}
		\centering
		\includegraphics[width=\linewidth]{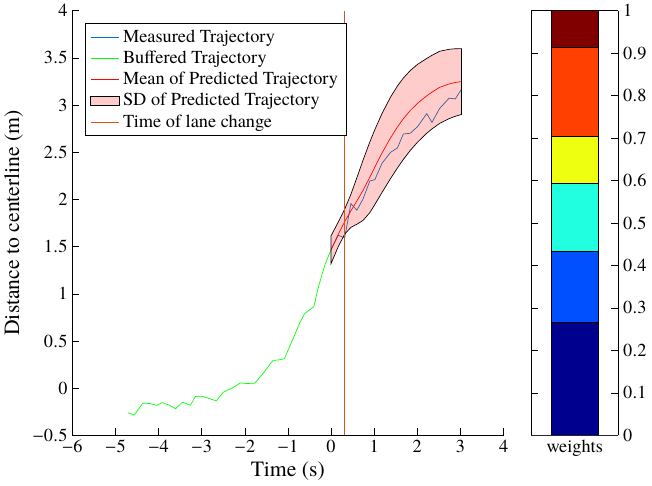}
		\caption{Predicted trajectory 0.32s before the lane change.}
		\label{fig_trajectory_prediction_0_32}
	\end{subfigure}
	\caption{Predicted trajectories (colored red) at two different time points before a lane change left. The vehicle's true motion is plotted in green for past states and in blue for future observations. The lane change event is indicated by a vertical red line. Bar plots depict the individual weights of each prototype in the ensemble model.}
	\label{fig_trajectory_prediction}
\end{figure*}

After successful driver intention recognition, the future trajectory of a vehicle is predicted according to the estimated maneuver. In \cite{Augustin.2018} the potential of prototype trajectories for both longitudinal and lateral motion prediction was demonstrated. In case of detected lane changes, the best matching prototype of the respective maneuver type, \gls{lcl} or \gls{lcr}, was selected as an estimate for future lateral motion. Here, we extend this idea by formulating a Gaussian mixture model combining and weighting the predictions of all prototype trajectories of the estimated maneuver type. The predicted lateral course can be calculated as a weighted sum of the prototype means:
\begin{equation} \label{eq_mu}
\mu(t) = \sum_{m} w_m \hat{\mu}_m(t)
\end{equation}
\noindent and the combined variance can be computed by:
\begin{eqnarray} \label{eq_sigma}
\sigma^2(t) & = &  \sum_{m} w_m \left( \hat{\sigma}^2_m(t) + \hat{\mu}^2_m(t) \right) - \mu^2(t)
\end{eqnarray}
\noindent The final values of mean and variance are hold when individual prototypes are evaluated outside their definition range.
\begin{eqnarray}
\hat{\mu}_m(t) & = &
\begin{cases}
\mu_m(\tau_m + t), & \tau_m + t < T_m\\
\mu_m(T_m), & \text{otherwise}
\end{cases} \\
\hat{\sigma}^2_m(t) & = &
\begin{cases}
\sigma^2_m(\tau_m + t), & \tau_m + t < T_m\\
\sigma^2_m(T_m), & \text{otherwise}
\end{cases}.
\end{eqnarray}
\noindent The weighting factors, $w_m$, are derived from the ratio of inversed Mahalanobis distance of each prototype representing a motion pattern of the detected maneuver to the partial trajectory:
\begin{equation}
w_m = \frac{\frac{1}{\bm{\Delta}(\bm{d}_p, \bm{\mu}_m)}}{\sum_{m}\frac{1}{\bm{\Delta}(\bm{d}_p, \bm{\mu}_m)}}.
\end{equation}
\noindent Figure \ref{fig_trajectory_prediction} depicts the mean and variance values of the prototype ensemble for different prediction horizons.

\noindent The longitudinal motion during a lane change maneuver can be described by a (nearly) constant acceleration model \cite{Augustin.2018}, also referred to as discrete Wiener process acceleration model \cite{bar2004estimation}. It is given by:
\begin{align}
\begin{bmatrix}
s_{k+1} \\
v_{s_{k+1}} \\
a_{s_{k+1}}
\end{bmatrix} 
=
\begin{bmatrix}
1 & T & \frac{1}{2}T^2\\
0 & 1 & T\\
0 & 0 & 1
\end{bmatrix}
\begin{bmatrix}
s_k\\
v_{s_k}\\
\overline{a}
\end{bmatrix}
+
\begin{bmatrix}
\frac{1}{2}T^2\\
T\\
1
\end{bmatrix}
\omega_a
\end{align}
with
\begin{equation}
\omega_a = \mathcal{N}(0, \sigma^2_{\overline{a}} )
\end{equation}
\noindent In \cite{Augustin.2018} the constant acceleration value is calculated averaging the longitudinal accelerations of all trajectories comprised in the cluster, which is the best fit to the vehicles' partial trajectory. $\sigma_{\overline{a}}^2$ is the related variance. Using the  Gaussian mixture model, the new values of $\overline{a}$ and $\sigma_{\overline{a}}^2$ can be computed via Equations \ref{eq_mu} and \ref{eq_sigma}, respectively.

\subsection{Smooth Transition to Prototype-based Trajectory Predictions}

The future lateral motion course, predicted based on the previously proposed model of mixed cluster representatives, does not necessarily pass through the currently observed state of the vehicle. Therefore, the transition from current lateral offset to future values is generally discontinuous. To solve this problem, we adapt all cluster prototypes that contribute to the predicted trajectory within the interval $[\tau_m, \tau_m+1 \, \mathrm{s}]$. This ensures continuity in lateral position, velocity, and acceleration also in the near future. The adaptations are based on transformation of the lateral position course into B-splines. B-splines and their basis functions have favorable characteristics allowing adaptations in limited scope (local support characteristic)  and guaranteeing continuity of a function and their derivatives of certain degree (continuity characteristic) \cite{piegl.2012}.\\
B-splines are piecewise polynomial curves of degree $d$ defined on an interval $[a, b]$. They are represented by a knot vector $\bm{t} = [t_0, t_1, ... , t_m]$ with $m + 1$ elements and a set of $n + 1$ control points $\bm{P} = \{\bm{P}_0,...,\bm{P}_n\}$. The relation $n + 1 = m + d$ is always valid. \\
\noindent B-Spline curves are represented by 
\begin{equation}
\textbf{C}(t) = \sum_{i=0}^{n} \bm{P}_i N_{i,d}(t) 
\end{equation}
where the basis functions $N_{i,d}(t)$ are recursively defined as
\begin{eqnarray}
N_{i,0}(t) & = &
\begin{cases}
1, & t \in [t_i,t_{i+1})\\
0, & \text{otherwise}
\end{cases}\\
N_{i,d}(t) & = & \frac{t - t_i}{t_{i+d} - t_i} N_{i,d-1}(t) + \frac{t_{i+d+1} - t}{t_{i+d+1} - t_{i+1}} N_{i+1,d-1}(t).
\end{eqnarray}
\noindent Initially, a cluster's polynomial representation $f(t) = \textbf{a}_0 + \textbf{a}_1 t + ... + \textbf{a}_n t^n$ is converted into a B-spline curve by scaling the polynomial to the interval $[0,1]$ and calculating the control points \cite{marsh2006applied} by 
\begin{equation}
\textbf{P}_i = \sum_{j=0}^{i} \frac{\binom{i}{j}}{\binom{d}{j}}\textbf{a}_j.
\end{equation}
\noindent Due to the local support characteristic, the control point $\bm{P}_i$ only affects the interval $[t_i,t_{i+p+1})$. We use this property to modify the trajectory within the first second of prediction by adding three knots at the interval boundaries, respectively. $\bm{C}(t)$ is $d-k = 2$ times continuously differentiable at these breakpoints, with multiplicity $k = 3$ corresponding to the number of identical knots in the knot vector. The insertion of a new knot $\hat{t} \in [t_s, t_{s+1})$ causes a new control point. As stated in \cite{marsh2006applied}, the elements of the control point set are updated by 
\begin{eqnarray}
\hat{\textbf{P}}_i & = & 
\begin{cases}
\textbf{P}_i, & 0 \leq i \leq s-d\\
(1-\alpha_i) \textbf{P}_{i-1} + \alpha_i \textbf{P}_i, & s-d+1 \leq i \leq s\\
\textbf{P}_{i-1}, & s+1 \leq i \leq n+1
\end{cases}\\
\alpha_i & = & \frac{\hat{t} - t_i}{t_{i+d} - t_i}. 
\end{eqnarray}
\noindent Within the first prediction second only a limited number of basis functions are unequal to zero. Due to  knot insertion the non-zero basis functions are $N_{4, 5}$, $N_{5, 5}$ and $N_{6, 5}$. The control points of these basis functions need to be adjusted such that the B-spline meets the currently observed vehicle states which are position, velocity and acceleration.
\noindent According to \cite{marsh2006applied}, the r-th derivative of a B-spline is defined as  
\begin{equation}
\textbf{C}^{(r)}(t) = \sum_{i=0}^{n-r} \textbf{P}^{(r)}_i N^{(r)}_{i,d-r}(t)
\end{equation}
\noindent where
\begin{equation}
\textbf{P}^{(r)}_i = (d-r+1)\frac{\textbf{P}^{(r-1)}_{i+1}-\textbf{P}^{(r-1)}_{i}}{t_{i+d+1} - t_{i+r}}
\label{eqn:bsplineder}
\end{equation}
\noindent with $\textbf{P}^0_i = \textbf{P}_i$. The knot vector is reduced to $t_r, ..., t_{m-r}$.
This yields the following equation system with three undefined variables $\textbf{P}_4$, $\textbf{P}_5$, and $\textbf{P}_6$:
\begin{eqnarray}
x & = & \textbf{P}_4 N_{4, 5}(\tau_m) + \textbf{P}_5 N_{5, 5}(\tau_m) + \textbf{P}_6 N_{6, 5}(\tau_m)
\label{eqn:bsplineeqnsystem1}\\
x' & = & \textbf{P}^{(1)}_4 N^{(1)}_{4, 4}(\tau_m) + \textbf{P}^{(1)}_5 N^{(1)}_{5, 4}(\tau_m)
\label{eqn:bsplineeqnsystem2}\\
x'' & = & \textbf{P}^{(2)}_4 N^{(2)}_{4, 3}(\tau_m).
\label{eqn:bsplineeqnsystem3}
\end{eqnarray}
\noindent The derivatives can be calculated via Equation \ref{eqn:bsplineder}.
The vehicle's lateral velocity and acceleration are estimated via derivation of a third-order polynomial approximation of the partial trajectory and inserted into Equations \ref{eqn:bsplineeqnsystem1}, \ref{eqn:bsplineeqnsystem2}, and \ref{eqn:bsplineeqnsystem3} to solve for the values of the required control points.


\section{Results} \label{results}
For evaluation and comparison of proposed approaches for maneuver estimation and subsequent trajectory prediction we use the data set proposed in \cite{Augustin.2018}. Aggregating hours of real highway footage, the whole data set comprises a total number of $434$ lane change maneuvers of traffic participants of which $156$ are lane changes to the left and $278$ to the right neighbor lane, respectively.
Data is divided with a ratio of $70/30$ into training and test set. During data division it is specially requested that ratio of maneuver types is equal in both fractions of the data set.	
In terms of maneuver classification we compute performance measures juxtaposing \gls{gda} and boosting decision trees. As the distribution of classes is highly unequal, with lane keeping clearly outnumbering lane change maneuvers, balanced versions of performance measures are applied.
The advances in trajectory prediction, namely combining prototype trajectories in a Gaussian mixture model and B-spline based trajectory smoothing, are compared to the previously implemented approach \cite{Augustin.2018}, which utilized solely the best matching cluster representative for future motion prediction.
\subsection{Evaluation of Maneuver Detection} \label{eval_maneuver_detection}
Due to the highly unequal class incidences, balanced measures \cite{bahram2016combined} must be applied for classification performance evaluation, whenever a quality criterion combines actual positive and actual negative portions of the regarded class (e.g. precision):
\begin{eqnarray}
\mathrm{TPR} & = & \frac{\mathrm{TP}}{\mathrm{TP}+\mathrm{FN}} = 1 - \mathrm{FNR}\\[1mm]
\mathrm{PRC} & = & \frac{\mathrm{TPR}}{\mathrm{TPR}+\mathrm{FPR}}\\[1mm]
\mathrm{F}_1 & = & \frac{2 \cdot \mathrm{PRC} \cdot \mathrm{TPR}}{\mathrm{PRC} + \mathrm{TPR}}
\end{eqnarray}
\noindent The True Positive Rate $\mathrm{TPR}$, also denoted recall, states how likely a maneuver will be predicted by the respective approach. The precision $\mathrm{PRC}$ indicates how likely a predicted lane change will actually happen. The $\mathrm{F}_1$ score is the harmonic mean of both precision and recall. Those values are computed on the whole test data set. Table \ref{tab} shows the classification performance for the proposed methods, Quadratic Gaussian Discriminant Analysis and Boosted Decision Trees, using two different sets of features, respectively. As a reference, to investigate the influence of features constructed from cluster prototypes, both methods are evaluated utilizing a two-dimensional feature vector $\bm{f}=[d,\dot{d}]^T$ comprising lateral offset, $d$, and lateral velocity, $\dot{d}$, of a vehicle in Frenet coordinates. In case of \gls{gda} we constructed two features utilizing distance values to the best matching prototypes by:
\begin{align}
f_3 & = \bm{\Delta}_{p,\mathrm{LCR}} - \bm{\Delta}_{p,\mathrm{LCL}} \\
f_4 & = \bm{\Delta}_{v,\mathrm{LCR}} - \bm{\Delta}_{v,\mathrm{LCL}}
\end{align}
\noindent Thereby, the dimension of the feature vector can be kept low while achieving good prediction results. For \gls{bdt} classifier we expand the feature vector to six dimensions $\bm{f}=[d, \dot{d}, \bm{\Delta}_{p,\mathrm{LCR}}, \bm{\Delta}_{p,\mathrm{LCL}}, \bm{\Delta}_{v,\mathrm{LCR}}, \bm{\Delta}_{v,\mathrm{LCL}}]^T$, allowing the ensemble model to learn mutual dependencies of features. Application of this feature vector for \gls{gda} worsens prediction quality, most likely because model assumptions are violated.
For the sake of comparison all classifiers were trained allowing a miss rate for the lane keeping maneuver of $MCR_{LK} = 0.11$. This is realized via adaptation of prior probabilities for \gls{gda} and initial weights of the data points for \gls{bdt}.
\begin{table}[h]
	\caption{Maneuver classification performance of Quadratic Gaussian Discriminant Analysis (GDA) and Boosted Decision Trees (\gls{bdt}) for different sets of features.} \label{tab}
	\begin{center}
		\def\arraystretch{1.5}
		\begin{tabular}{|c|c|c|c|c|c|c|c|}
			\hline
			Approach & Features & Maneuver & Miss Rate & Recall & Precision & $\mathrm{F}_1$-Score & Avg. Prediction Time \\
			\hline
			\hline
			\multirow{3}{*}{GDA} & \multirow{3}{*}{$d$, $\dot{d}$} & LCL & 0.18 & 0.82 & 0.95 & 0.880 & \multirow{3}{*}{1.54}\\
			\cline{3-7}
			& & LK & 0.11 & 0.89 & 0.81 & 0.849 & \\
			\cline{3-7}
			& & LCR & 0.23 & 0.77 & 0.94 & 0.845 & \\
			\hline
			\hline
			\multirow{3}{*}{GDA} & $d$, $\dot{d}$, & LCL & 0.16 & 0.84 & 0.96 & 0.894 & \multirow{3}{*}{1.68}\\
			\cline{3-7}
			& $\bm{\Delta}_{p,\mathrm{LCR}} - \bm{\Delta}_{p,\mathrm{LCL}}$, & LK & 0.11 & 0.89 & 0.82 & 0.855 & \\
			\cline{3-7}
			& $\bm{\Delta}_{v,\mathrm{LCR}} - \bm{\Delta}_{v,\mathrm{LCL}}$ & LCR & 0.22 & 0.78 & 0.93 & 0.851 & \\
			\hline
			\hline
			\multirow{3}{*}{\gls{bdt}} & \multirow{3}{*}{$d$, $\dot{d}$} & LCL & 0.13 & 0.87 & 0.96 & 0.915 & \multirow{3}{*}{1.61}\\
			\cline{3-7}
			& & LK & 0.11 & 0.89 & 0.85 & 0.870 & \\
			\cline{3-7}
			& & LCR & 0.18 & 0.82 & 0.94 & 0.872 & \\
			\hline
			\hline
			\multirow{3}{*}{\gls{bdt}} & $d$, $\dot{d}$, & LCL & 0.11 & 0.89 & 0.96 & 0.921 & \multirow{3}{*}{1.67}\\
			\cline{3-7}
			& $\bm{\Delta}_{p,\mathrm{LCR}}$, $\bm{\Delta}_{p,\mathrm{LCL}}$, & LK & 0.11 & 0.89 & 0.85 & 0.874 & \\
			\cline{3-7}
			& $\bm{\Delta}_{v,\mathrm{LCR}}$, $\bm{\Delta}_{v,\mathrm{LCL}}$ & LCR & 0.18 & 0.82 & 0.94 & 0.877 & \\
			\hline
		\end{tabular}
	\end{center}
\end{table}
\subsection{Discussion of Maneuver Detection Performance} \label{discuss_maneuver_detection}
First, the influence of features constructed from cluster prototypes on the classification performance is investigated for both learning methods.
\gls{gda} strongly benefits from additional information provided by prototype trajectories. The average prediction time can be increased by $0.14 \, \mathrm{s}$ to a total of $1.67 \, \mathrm{s}$. The performance measures show improvements, as well. The increase in average prediction time of $0.06 \, \mathrm{s}$ for the \gls{bdt} method by adding prototype-derived features is comparatively small, but again, the prediction quality measures denote performance improvements.\\
For the two-dimensional feature vector $\bm{f} = [d, \dot{d}]^T$ \gls{bdt} performs significantly better reaching an improved prediction time and higher $\mathrm{F}_1$-scores. Whereas the precision values are on equal level, the major improvements are reported by increased recall values for lane change maneuvers, which means that actual lane changes are more likely to be detected. Incorporating information provided by motion pattern recognition, both \gls{gda} and \gls{bdt} achieve equal average prediction horizons of roughly $1.7 \, \mathrm{s}$, while, again, \gls{bdt} method shows better values for performance measures. The optimal average prediction horizon according to the automatically labeled data can be calculated to $2 \, \mathrm{s}$. 
\subsection{Evaluation of Motion Prediction} \label{eval_motion_predicion}
\begin{figure*}[t]
	\begin{subfigure}{0.45\linewidth}
		\centering
		\includegraphics[width=0.9\linewidth]{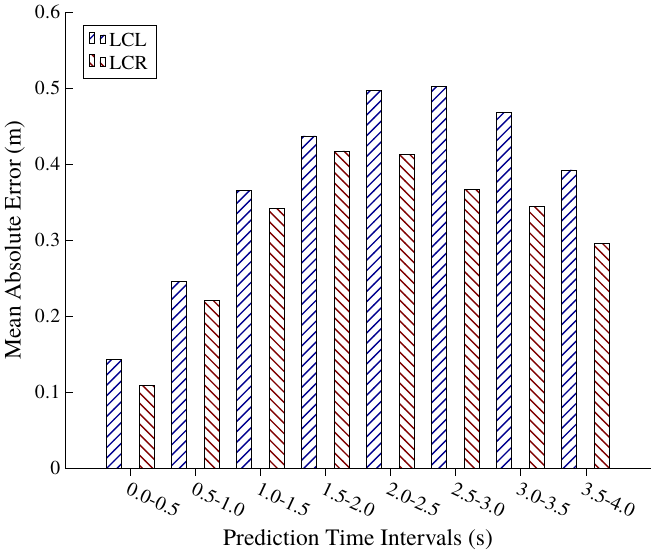} 
		\caption{Lateral errors of the reference approach.}
	\end{subfigure}\vspace{4mm}
	\begin{subfigure}{0.45\linewidth}
		\centering
		\includegraphics[width=0.9\linewidth]{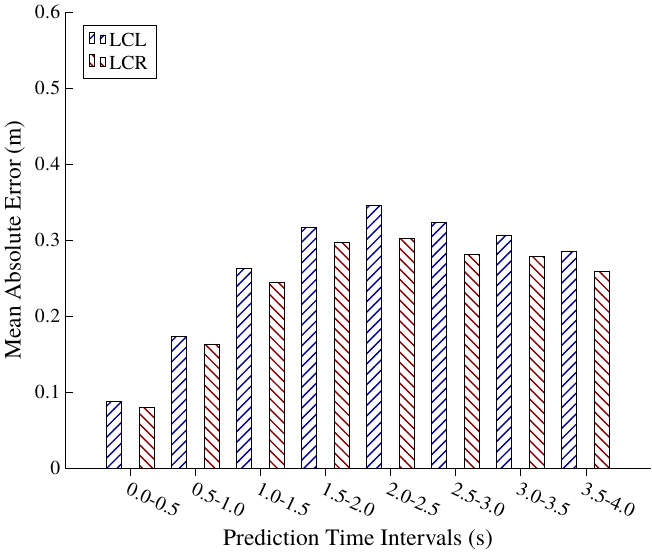}
		\caption{Lateral errors of the proposed approach.}
	\end{subfigure}
	\begin{subfigure}{0.45\linewidth}
		\centering
		\includegraphics[width=0.9\linewidth]{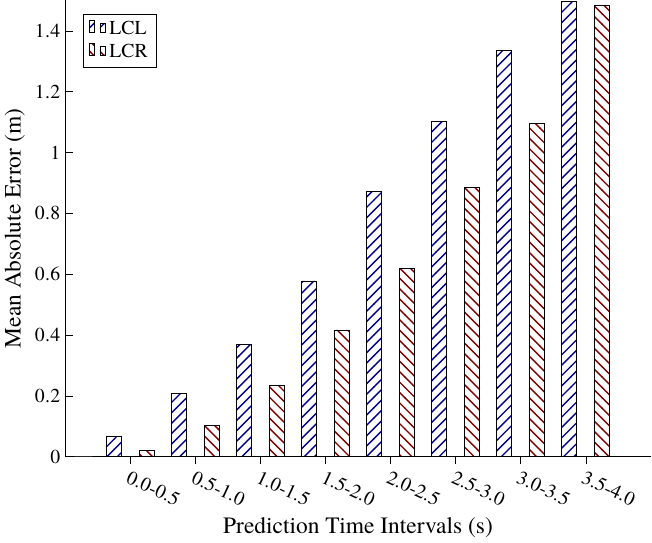} 
		\caption{Longitudinal errors of the reference approach.}
	\end{subfigure}%
	\begin{subfigure}{0.45\linewidth}
		\centering
		\includegraphics[width=0.9\linewidth]{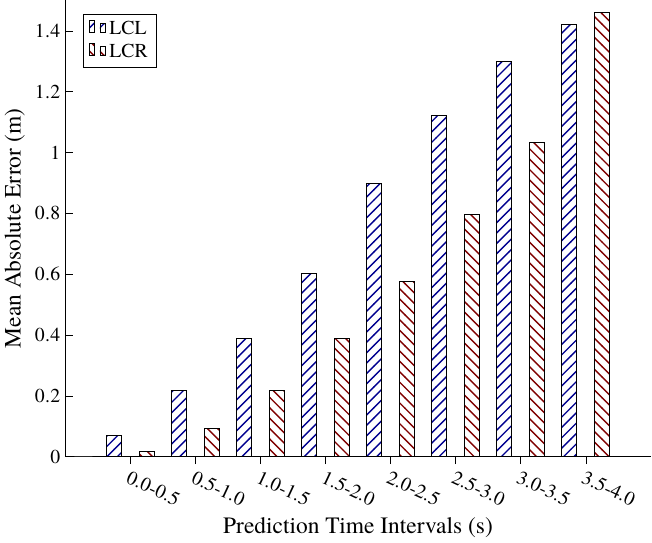}
		\caption{Longitudinal errors of the proposed approach.}
	\end{subfigure}
	\caption{Mean absolute errors between the traffic participants' predicted and actual positions for various prediction horizons.}
	\label{fig_error}
\end{figure*}

An advantage of prototype-based maneuver detection lies in the usability of cluster prototypes for the future motion prediction of traffic participants. Here, the extended approach for trajectory prediction, as described in Section \ref{motion_prediction}, is evaluated and compared to the original version presented in \cite{Augustin.2018}.\\
Figure \ref{fig_error} shows the mean absolute lateral and longitudinal errors between traffic participants' predicted and actual positions for different prediction horizons. The evaluation results for the reference approach is illustrated on left, while the new motion prediction results are depicted on the right side of the figure.
The prediction horizon $[0 \, \mathrm{s}, 4\, \mathrm{s}]$, which is plotted on the horizontal axis, is subdivided into intervals of $\Delta t = 0.5 \, \mathrm{s}$.
For the $n$-th prediction interval, the mean absolute position error is calculated by
\begin{equation}
\epsilon_{n} = \frac{1}{N_n} \cdot \sum_{\tau \in [n \Delta T, (n+1) \Delta T)} | \bm{x}_t - \hat{\bm{x}}_{(t|t-\tau)} |
\end{equation}
\noindent where $N_n$ is the total number of summands, $\bm{x}_t = [s_t, d_t]$ denotes the actual observation at time $t$ and $\hat{\bm{x}}_{(t|t-\tau)}$ is the prediction of vehicle state at time $t$ calculated at time $t - \tau$. The maximum evaluation time is limited by the length of the prototype ensemble used for motion prediction and capped to a value of $4 \, \mathrm{s}$.
The evaluation starts when the correct lane change maneuver is detected continuously and ends at the point in time when the vehicle crosses the lane marking.
\subsection{\textbf{Discussion of Motion Prediction Performance}} \label{discuss_motion_predicion}
\begin{figure*}[t]
	\centering
	\includegraphics[width=0.375\linewidth]{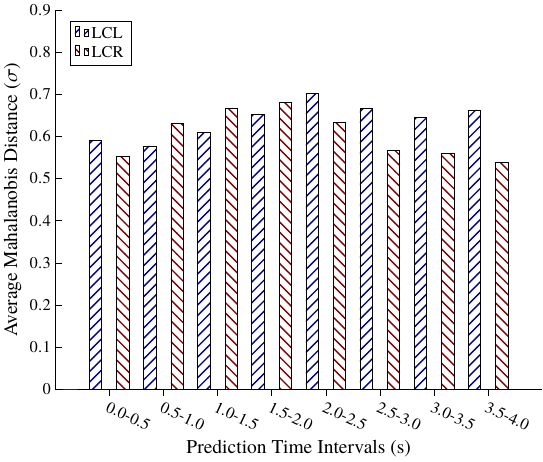} 
	\caption{Mean lateral Mahalanobis distance between the traffic participants' predicted and actual positions for various prediction horizons.}
	\label{fig_mahal_error}
\end{figure*}

For longitudinal trajectory prediction during lane change maneuvers both approaches perform equally well. Comparing both lane change types, the errors are generally larger in case of \gls{lcl} maneuvers. The reason for this lies in the larger diversity of longitudinal accelerations during lane changes to the left.\\
The performance in lateral motion prediction is significantly improved by the extensions presented in Section \ref{motion_prediction}. For small prediction horizons, the absolute position errors correspond approximately to the measurement accuracy. For a prediction of horizon of $2 - 4 \, \mathrm{s}$, the absolute lateral error is steady with an approximate value of $0.3 \, \mathrm{m}$. 		
In case of  the reference approach, error peaks for medium prediction horizons can be observed. These peaks result from incorrect selections of prototype trajectories in an early stage of the detected maneuver. At the beginning many cluster prototypes are similar so that wrong prototype selection is quite likely. The combination of prototypes in a Gaussian mixture model can compensate this issue.\\
Figure \ref{fig_mahal_error} plots the average Mahalanobis distance between traffic participants' predicted and actual lateral offsets. The upper bound for all considered prediction horizons is $0.7 \, \sigma$, which indicates that the  variance of the prototype ensemble, which is calculated utilizing Equation \ref{eq_sigma}, is a good estimate of prediction uncertainty.

\section{Conclusion and Outlook}

In this paper, a statistical approach is proposed, which successfully utilizes a set of prototypical lane change trajectories to realize both early maneuver detection and uncertainty-aware trajectory prediction for traffic participants. Generation of prototype trajectories from real traffic data is accomplished by \glsdesc{ahc}, optimizing mutual cluster configurations and constraining cohesion during the clustering process \cite{Augustin.2018}. The main contributions of this paper exist in targeted adaptations of prediction methods to improve utilization of motion patterns and thus enhancing prediction results. For maneuver recognition, we implemented \glsdesc{bdt} classification increasing the detection rate of lane change maneuvers compared to previously applied \glsdesc{gda}. The future trajectory is predicted according to typical realizations of the estimated maneuver. We introduced a mixture model of cluster prototypes and demonstrated increased accuracy for lateral motion prediction. Additionally, a B-splines based adaptation technique is described guaranteeing continuity during transition from actually observed to predicted vehicle states.
In future work, the prediction framework is extended by taking interaction between traffic participants into account to further increase the prediction horizon and lower misclassification rates. 

\bibliographystyle{unsrt}
\bibliography{literatur_autoreg_I}


\end{document}